\documentclass[11pt]{article}
\usepackage{coling2020}
\usepackage{times}
\usepackage{url}
\usepackage{latexsym}
\usepackage{graphicx}
\usepackage{booktabs}

\colingfinalcopy 

\title{Unequal Representations: Analyzing Intersectional Biases in Word Embeddings Using Representational Similarity Analysis}

\author{Michael A. Lepori \\
        Department of Computer Science\\ Johns Hopkins University\\
  {\tt mlepori19@gmail.com} }

\date{}

\begin{document}
\maketitle
\begin{abstract}
We present a new approach for detecting human-like social biases in word embeddings using representational similarity analysis. Specifically, we probe contextualized and non-contextualized embeddings for evidence of intersectional biases against Black women. We show that these embeddings represent Black women as simultaneously less feminine than White women, and less Black than Black men. This finding aligns with intersectionality theory, which argues that multiple identity categories (such as race or sex) layer on top of each other in order to create unique modes of discrimination that are not shared by any individual category.
\end{abstract}

\section{Introduction}
\label{intro}

%
%
\blfootnote{
    %
    %
    %
    %
    %
    %
    \hspace{-0.65cm}  
    This work is licensed under a Creative Commons 
    Attribution 4.0 International License.
    License details:
    \url{http://creativecommons.org/licenses/by/4.0/}.
}

 Word embeddings are ubiquitous components of modern natural language processing systems, and have enabled performance increases on a wide variety of NLP tasks \cite{devlin2019bert,peters2018deep,pennington2014glove}. They provide vector representations of words, and are trained on a large corpus of text in order to capture the distributional semantics of lexical items. Many studies have shown that these embeddings also encode a variety of dangerous social biases, as these biases are present in naturally-occurring English text (see \newcite{mehrabi2019survey} for a survey of such work).
 
 The present work studies \textit{intersectional biases} in both contextualized and non-contextualized word embeddings. The theory of intersectionality states that multiple different aspects of a person's identity often combine to create unique modes of discrimination \cite{crenshaw1989demarginalizing,crenshaw1990mapping}. For example, specific biases against Black women are not necessarily shared by either White women or Black men. Furthermore, discourses on discrimination and public health typically focus on sexism \textit{or} racism, each implicitly excluding the other \cite{crenshaw1989demarginalizing,bowleg2012problem}.
 
It is further argued that these discourses are focused on the most privileged members of a group \cite{crenshaw1989demarginalizing}. Thus, we might expect that the ostensibly race-neutral category `Woman' is implicitly White, and that the ostensibly non-gendered category `Black' is implicitly male. Using representational similarity analysis (RSA; \newcite{kriegeskorte2008representational}), we find that both non-contextualized GloVe embeddings \cite{pennington2014glove} and contextualized BERT embeddings \cite{devlin2019bert} display exactly these biases. We also find that the BERT embeddings of the names of Black women are semantically impoverished, compared to the embeddings of the names of White women.
 
 \paragraph{Related Work:} 
Many recent studies have sought to understand social biases in NLP systems. Notably, \newcite{caliskan2017semantics} showed that the human biases revealed through the Implicit Association Test \cite{greenwald1998measuring} are also present in word embeddings, prompting other studies that seek to detect bias in word embeddings \cite{kurita2019measuring,garg2018word}. Other studies have attempted to detect bias in other NLP systems \cite{rudinger2018gender}, to automatically discover and measure intersectional biases in word embeddings \cite{guo2020detecting}, to analyze the distributions of gendered words in training sets \cite{zhao-etal-2019-gender,rudinger2017social}, and to automatically mitigate the effects of social biases on NLP systems \cite{sun2019mitigating}. Another study has produced empirical evidence for the claims of intersectionality theory by leveraging distributional representations \cite{herbelot2012distributional}. 

 Perhaps most relevant, some prior work has attempted to understand whether intersectional differences are present in contextualized word embeddings by trying to detect whether these embeddings encoded intersectional stereotypes \cite{may2019measuring,tan2019assessing}. However, detecting these stereotypes is only a proxy for detecting intersectional biases in embeddings: An embedding might encode intersectional differences without encoding more complex interesectional stereotypes. This work presents a method that leverages representational similarity analysis to directly probe word embeddings for intersectional biases.\footnote{Code and data found at \url{https://github.com/mlepori1/Unequal_Representations}.}

\section{Methods: Representational Similarity Analysis}
\label{rsaSect}

Representational similarity analysis (RSA) is a technique first used in cognitive neuroscience for analyzing distributed activity patterns in the brain \cite{kriegeskorte2008representational}, but recent work has used it to analyze artificial neural systems \cite{leporiRSA,chrupala-2019-symbolic,chrupala2019correlating,abnar-etal-2019-blackbox,bouchacourt-baroni-2018-agents}. RSA analyzes the representational geometry of a system, which is defined by the pairwise dissimilarities between representations of a set of stimuli. Specifically, we use RSA to compare the representational geometry of the word embeddings under study to the representational geometries of well-understood \textit{hypothesis models}. If the representational geometries are similar, you can infer that the word embeddings represent the particular information expressed by a hypothesis model.

\paragraph{Applying RSA:}
The approach proceeds as follows: First, we create a corpus that consists of three types of items: group 1 items, $G_1$, group 2 items, $G_2$, and concept items, $C$. Next, we define a reference model $M_{Ref}$, which consists of the set of embeddings that we are interested in. Then, we define two hypothesis models, $M_{Hyp_1}$ and $M_{Hyp_2}$. $M_{Hyp_1}$ instantiates the hypothesis that: `Group 1 is associated with the concept under study, while Group 2 is not'. Likewise $M_{Hyp_2}$ instantiates the hypothesis that `Group 2 is associated with the concept under study, while Group 1 is not'. We then draw a sample $c$, consisting of $n_1$ items from $G_1$, $n_2$ items from $G_2$, and $n_3$ items from $C$, and calculate the $(n_1 + n_2 + n_3) \times (n_1 + n_2 + n_3)$ representational geometries $R$ of each model using a dissimilarity metric $D$. In all analyses, $n_1 = n_2 = n_3 = 10$.
\begin{equation}
    R_{Ref} = D(M_{Ref}, c)
\end{equation}
\begin{equation}
    R_{Hyp_1} = D(M_{Hyp_1}, c)
\end{equation}
\begin{equation}
    R_{Hyp_2} = D(M_{Hyp_2}, c)
\end{equation}

We define $D = 1 - Spearman's\:\rho$ for $M_{Ref}$. For $M_{Hyp_1}$, the dissimilarity metric is equal to 1 if one item is from $G_2$ and the other item is from $G_1$ or $C$, and 0 otherwise. Likewise, the dissimilarity metric for $M_{Hyp_2}$ is equal to 1 is one item is from $G_1$ and the other item is from $G_2$ or $C$, and 0 otherwise. Clearly, these hypothesis models are not expected to provide a very good fit to the representational geometry of a set of word embeddings. However, if one hypothesis model exhibits a significantly \textit{better} fit than the other, then there is evidence for a problematic bias.

Finally, we calculate the similarity, $s$, between the representational geometries of our hypothesis models and our reference model using a similarity metric, $sim$. Because the $R$ matrices are symmetric, $sim$ only operates on the upper triangle of each $R$ matrix. We define $sim = Spearman's\:\rho$.
\begin{equation}
    s_{Hyp_1} = sim(R_{Ref}, R_{Hyp_1})
\end{equation}
\begin{equation}
    s_{Hyp_2} = sim(R_{Ref}, R_{Hyp_2})
\end{equation}

\begin{figure}[h!]
 \centering
  \resizebox{.8\textwidth}{!}{
  \includegraphics[width=15.5cm]{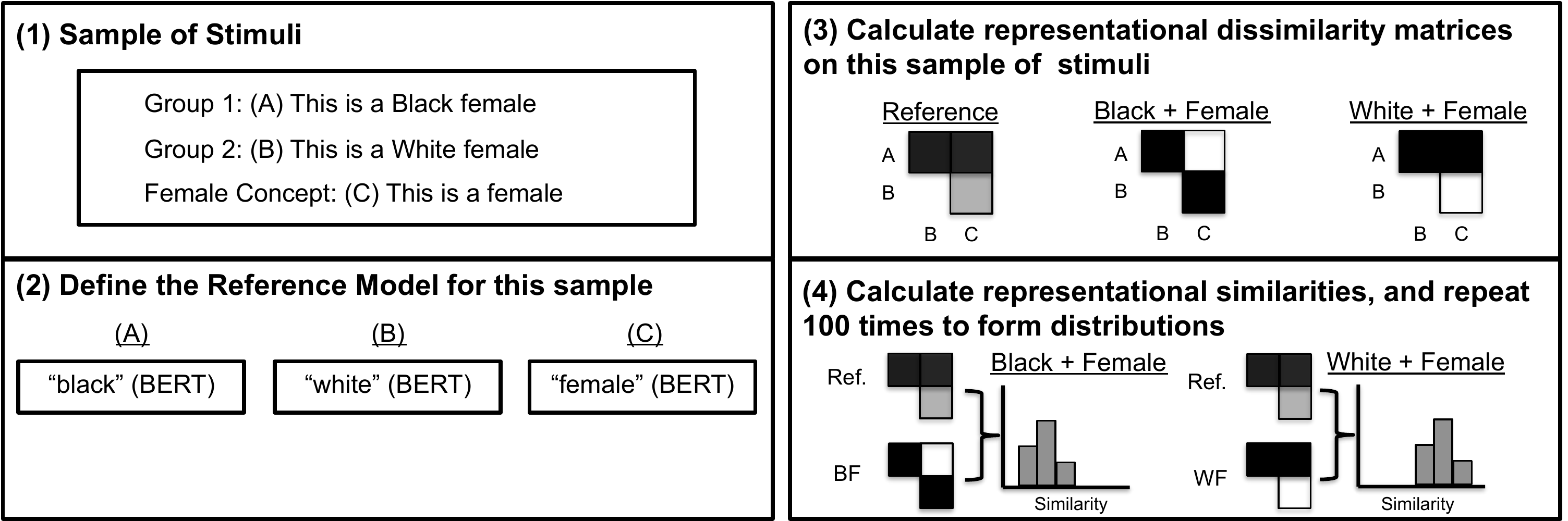}
  }
    \caption{A summary of our approach, applied to BERT embeddings.}
    \label{fig:summaryfig}
    
\end{figure}

We then repeat the process on 100 samples from our corpus in order to create two 100-length vectors of representational similarities, $S_{Hyp_1}$ and $S_{Hyp_2}$. We can then apply a nonparametric sign test to the difference of these vectors, $S_{Hyp_1} - S_{Hyp_2}$, in order to test for a consistent difference between measurements of $s_{Hyp_1}$ and $s_{Hyp_2}$. See Figure~\ref{fig:summaryfig} for a visual summary of this approach. 

\section{Data}
\label{data}
In order to perform the analyses described in Section~\ref{rsaSect}, we need to create sets of group items for all of the groups under study, as well as concept items for the identity concepts under study. For analyses involving GloVe embeddings, these glossaries can only contain single-word entries. For analyses involving BERT embeddings, these entries can be full sentences. See Appendix~\ref{data_app} for more details about all of the following datasets.
\subsection{Single-Word Sets}
\paragraph{Group Items:}For our single-word group items for the Black women, Black men, White women, and White men groups, we use the dataset curated in \newcite{sweeney2013discrimination}. This dataset includes names commonly associated with all of these identity groups. We rely on name data because identity categories like \textit{Black woman} do not have single word terms \cite{may2019measuring}. See Section~\ref{limits} for a discussion of the limitations of using name data.

\paragraph{Concept Items:} For our single-word, ostensibly race-neutral concept items that are associated with the female identity category, we combine the datasets from \newcite{nosek2002harvesting} and \newcite{nosek2002math}.
The United States Census Bureau defines the Black/African American identity category as ``all individuals who identify with one or more nationalities or ethnic groups originating in any of the black racial groups of Africa" \cite{census_bureau_2020}. They proceed to non-exhaustively list several nationalities that meet this definition. All 10 single-word country names that correspond to the nationalities listed are included in our set of single-word, ostensibly gender-neutral concept items that are associated with the Black identity category. We also include the word \textit{Black}, as well as the word \textit{Africa}, in keeping with the Census Bureau's definition. See Section~\ref{limits} for a discussion of the limitations of this approach.

\subsection{Multi-Word Sets}
We leverage the semantically bleached sentence templates introduced by \newcite{may2019measuring} in order to generate our sentence data. These sentences ``make heavy use of deixis and are designed
to convey little specific meaning beyond that of the
terms inserted into them". Some examples include: ``This is a $<$word/phrase$>$" and ``The $<$word/phrase$>$ is here".
\paragraph{Group Items:} For the Black female group, we fill in the semantically-bleached templates with either the word \textit{Black} concatenated with either \textit{woman, women, female, females, girl,} or \textit{girls} (such that the template sentence exhibits proper number agreement). For example, the dataset contains the sentence: \textit{The Black woman is here}. Similarly for the White female group. These sentences are also used by \newcite{may2019measuring}. Finally, for the Black male group, we fill in the template with the word \textit{Black} and either \textit{man, men, male, males, boy,} or \textit{boys}. 
\paragraph{Concept Items:} Our female concept items are chosen to be ostensibly race-neutral. Thus, we fill in the template sentences with one of the following words: \textit{woman, women, female, females, girl,} or \textit{girls}. Our Black concept items are chosen to be ostensibly gender-neutral. Thus we fill in the template sentences with either \textit{Black person} or \textit{Black people}.

\section{Experiments and Results}
\subsection{GloVe Experiments}
\label{glove}
GloVe embeddings are non-contextualized word embeddings, and so they represent information from one word, ignoring the specific context in which they are found. These experiments seek to test whether RSA can be used to identify intersectional biases in GloVe embeddings. We use 300-dimensional embeddings trained on the Wikipedia~2014~+~Gigaword~5 corpus throughout this study.\footnote{Note: We have slightly revised all values in Table~\ref{resultsTable} pertaining to the Female Concept, due to a small data error in the original COLING 2020 publication. This did not impact any of our conclusions. See Appendix~\ref{data_app} for further discussion.}

\paragraph{Black Female Name Embedding Check}\label{bfn_sanity}First, we ensure that the GloVe embeddings of the names of Black women contain sufficient semantic detail, such that more in-depth experiments can be performed. Our concept items are a set of words associated with the female identity category. Our Group~1 items are a set of proper names associated with Black women, and our Group~2 items are a set of proper names associated with Black men. See Section~\ref{data} for details about these sets. We would expect that the reference model will display greater representational similarity to the hypothesis model that associates female attributes to Black female names. From Table~\ref{resultsTable} we see that this happens. 

\paragraph{Female Concept} We investigate whether the GloVe embeddings of the names of Black women encode the `female' concept less than the embeddings of the names of White women. Our concept items are a set of words associated with the female identity category. Our Group~1 items are a set of proper names associated with Black women, and our Group~2 items are a set of proper names associated with White women. See Section~\ref{data} for details about these sets. From Table~\ref{resultsTable}, we see that the reference model exhibits greater representational similarity to the hypothesis model that associates the concept items to White female names.

\paragraph{Black Concept} We investigate whether the GloVe embeddings of the names of Black women encode the `Black' concept less than the embeddings of the names of Black men. Our concept items are a set of words associated with the Black identity category. Our Group~1 items are a set of proper names associated with Black women, and our Group~2 items are a set of proper names associated with Black men. From Table~\ref{resultsTable}, we see that the reference model exhibits greater representational similarity to the hypothesis model that associates the concept items with Black male names.

\subsection{BERT Experiments}
BERT embeddings are contextualized word embeddings, and so they represent information from a word in context. This allows us to analyze group terms, using the semantically-bleached sentence data from \newcite{may2019measuring}, which are designed to convey little meaning beyond that of the group term that we insert into them. We also attempt to reproduce our findings using the single-word proper name data used in Section~\ref{glove}. We use the pretrained BERT-base-uncased model for all multi-word studies.
\paragraph{Black Female Name Embedding Check:} First, we use the pretrained BERT-base-cased model to attempt to reproduce our results from Section~\ref{glove}. However, we see from Table~\ref{resultsTable} that the reference model displays greater representational similarity to the hypothesis model that associates the female concept items to Black \textit{male} names. This indicates that the BERT's embeddings for Black female names are quite semantically impoverished, and do not reliably encode information about identity categories. Thus, we do not attempt to replicate our investigations of intersectional biases using BERT's embeddings of proper names. Notably, this effect is \textit{not} reproduced when analyzing White names, and thus demonstrates a problematic asymmetry between the BERT representations of White and Black female names.


\paragraph{Female Concept:} We investigate whether the BERT embeddings of the word \textit{Black} encode the `female' concept less than the embeddings of the word \textit{White}. Our concept items are a set of sentences containing words associated with the female identity category. Our Group~1 items are as a set of sentences referencing Black women, our Group~2 items are a set of sentences referencing White women. For each sentence in the Group sets, we extract the BERT embedding of the word \textit{Black} or \textit{White}, and for each sentence in the Concept set, we extract the embedding of the word associated with the female identity category (\textit{woman}, \textit{female}, etc.). See Section~\ref{data} for more details. From Table~\ref{resultsTable}, we see that the reference model exhibits greater representational similarity to the hypothesis model that associates the female concept items with White women.

\paragraph{Black Concept:} We investigate whether the BERT embeddings of words associated with the female identity category encode the `Black' concept less than the embeddings of words associated with the male identity category. Our concept items are a set of sentences containing the phrase \textit{Black person} or \textit{Black people}. Our Group~1 items are a set of sentences referencing Black women, our Group~2 items are a set of sentences referencing Black men. For each sentence in the Group sets, we extract the BERT embedding of the word associated with either the female or male identity category, and for each sentence in the Concept set, we extract the embedding of the word \textit{Black}. See Section~\ref{data} for more details. From Table~\ref{resultsTable}, we see that the reference model exhibits greater representational similarity to the hypothesis model that associates the Black concept items with Black men.

\begin{table*}[!h]
\centering
\resizebox{.8\textwidth}{!}{%
\begin{tabular}{c c c c c c c} 
 \toprule
 Embedding & Group 1 & Group 2 & Concept & Group 1 + Con & Group 2 + Con \\ [0.5ex] 
 \midrule
 GloVe & B F Names & B M Names & Female & \textbf{.375} & .176 \\ 
 GloVe & B F Names & W F Names & Female & .091 & \textbf{.669} \\
 GloVe & B F Names & B M Names & Black & .174 & \textbf{.298} \\
 BERT & B F Names & B M Names & Female & .116 & \textbf{.323} \\
 BERT & W F Names & W M Names & Female & \textbf{.373} & .020 \\
 BERT & B F Sent & W F Sent & Female & .050 & \textbf{.110} \\
 BERT & B F Sent & B M Sent & Black & .253 & \textbf{.289} \\
 \bottomrule
\end{tabular}
}
\caption{Results for all experiments. All differences statistically significant ($p < .001$)}
\label{resultsTable}
\end{table*}

 \section{Limitations}\label{limits} One notable limitation of this work is that we study artificial binary gender and race distinctions. Unfortunately, these binary distinctions exclude many individuals from our analysis. Future work should expand upon the methodological framework presented here in order to assess biases against groups that are not represented by these distinctions. Furthermore, \newcite{gaddis2017black} describes several limitations of relying on name data to represent racial categories. Perhaps most importantly, that work demonstrates that `Black names' are not one monolithic category, that perceptions of the Blackness of names are correlated with other confounding variables (such as education status), and that the associations between identity categories and proper names are only loosely reflected in real-world naming practices. Furthermore, we rely on an incomplete set of country names provided by the Census Bureau to generate concept sets for the Black identity category. We acknowledge that this excludes many nationalities from our analysis, and also acknowledge that any attempt to exhaustively list nations associated with the Black identity category is destined to fail. We hope that our results inspire future work that aspires to more robust coverage of the Black identity group. Finally, it is important to note that this method can only provide positive evidence: If this method does not detect intersectional biases for a particular system, that does not provide evidence that the system does not contain such biases.

\section{Conclusion}
Both contextualized and non-contextualized word embeddings learn to represent human-like biases. We apply RSA to the task of detecting these biases, and find that both GloVe embeddings and BERT embeddings reproduce (and thus perpetuate) the dual marginalization of Black women within both the Black community and female community, as predicted by the theory of intersectionality. Specifically, we showed that word embeddings represent the female identity category as implicitly White, and the Black identity category as implicitly masculine. To our knowledge, this is the first method that can probe word embeddings for these intersectional biases directly. Aside from addressing the issues discussed in Section~\ref{limits}, future work should investigate whether this method can be used to detect other sorts of intersectional biases, particularly those facing the LGBTQ community.

\bibliographystyle{coling}
\bibliography{coling2020}

\newpage
\appendix
\section{Data}
\label{data_app}

\subsection{Proper Names}

We use the proper name data found in \newcite{sweeney2013discrimination}. That work largely compiles these proper names from two previous studies, which rely on correlations between race and child names in birth records from the years 1974-1979 in Massachusetts \cite{bertrand2004emily}, and in birth records from 1961-2000 in California \cite{fryer2004causes}. We omit names that are polysemous. This occurs for four of the Black female names: \textit{Diamond}, \textit{Precious}, \textit{Kenya}, and \textit{Ebony}. One name (\textit{Trevon}) did not occur in the Wikipedia 2014 + Gigaword 5 corpus, and so it was also excluded from our analysis. We randomly omit names from the Black male, White female, and White male proper name sets so that all identity categories are represented by an equal number (13) of names. The list of names from \newcite{sweeney2013discrimination} is found below. Italicized names were omitted.

\paragraph{Black Female:} Aisha, Keisha, Latonya, Lakisha, Latoya, Tamika, Imani, Shanice, Aaliyah, Nia, Latanya, Latisha, Deja, \textit{Diamond}, \textit{Precious}, \textit{Kenya}, \textit{Ebony}
\paragraph{Black Male:} Darnell, Hakim, Jermaine, Kareem, Jamal, Leroy, Rasheed, DeShawn, DeAndre, Marquis, Terrell, Malik, Tyrone, \textit{Tremayne}, \textit{Trevon}
\paragraph{White Female:} Allison, Anne, Carrie, Emily, Jill, Laurie, Kristen, Meredith, Molly, Amy, Claire, Madeline, Emma, \textit{Katie}
\paragraph{White Male:} Brad, Brendan, Geoffrey, Greg, Brett, Jay, Matthew, Jake, Connor, Tanner, Wyatt, Cody, Dustin, \textit{Neil}

\subsection{Female Concept Words}
We combine the datasets of words associated with the female identity category from \newcite{nosek2002harvesting} and \newcite{nosek2002math}, which are two often-cited studies that leverage the implicit association test \cite{greenwald1998measuring}. Note that in the original COLING 2020 publication of this work, we had calculated our results with a dataset that replaces \textit{aunt} with \textit{niece} and omits \textit{grandmother}. For consistency with our cited work, we have revised our values using the dataset listed below. The list is as follows:
\paragraph{Female Concept Words:} female, woman, girl, sister, she, her, hers, daughter, aunt, mother, grandmother

\subsection{Black Concept Words}
Our Black concept words are derived from the definitions given by the United States Census Bureau \cite{census_bureau_2020}. The list is as follows:
\paragraph{Black Concept Words:} Africa, Black, Jamaica, Haiti, Nigeria, Ethiopia, Somalia, Ghana, Barbados, Kenya, Liberia, Bahamas

\subsection{Multi-Word Group Items}
Our group items leverage the semantically-bleached sentence templates of \newcite{may2019measuring}. We list some examples of group items for all identity groups studies.

\paragraph{Black Female} This is a Black female.; The Black woman is here.; They are Black girls.

\paragraph{White Female} This is a White female.; The White woman is here.; They are White girls.

\paragraph{Black Male} This is a Black male.; The Black man is here.; They are Black boys.

\paragraph{White Male} This is a White male.; The White man is here.; They are White boys.

\subsection{Multi-Word Female Concept Items}
Our multi-word female concept items leverage the semantically-bleached sentence templates of \newcite{may2019measuring}. We list some examples of Female concept items.

\paragraph{Female Concept Items} Females are people.; The girl is there.; Here is a woman.

\subsection{Multi-Word Black Concept Items}
Our multi-word Black concept items leverage the semantically-bleached sentence templates of \newcite{may2019measuring}. We list some examples of Black concept items.

\paragraph{Black Concept Items} Here is a Black person.; These are Black people.; Black people are people.

\end{document}